\documentclass{article}

\usepackage{PRIMEarxiv}

\usepackage[utf8]{inputenc} 
\usepackage[T1]{fontenc}    
\usepackage{hyperref}       
\usepackage{url}            
\usepackage{booktabs}       
\usepackage{amsfonts}       
\usepackage{nicefrac}       
\usepackage{microtype}      
\usepackage{lipsum}
\usepackage{fancyhdr}       
\usepackage{graphicx}       
\graphicspath{{media/}}     

\usepackage{amssymb}

\usepackage{mathtools}
\usepackage{algorithm}
\usepackage[noend]{algpseudocode}

\usepackage{enumitem}
\newlist{questions}{enumerate}{2}
\setlist[questions,1]{label=\textbf{RQ\arabic*.},ref=RQ\arabic*,leftmargin=*}
\setlist[questions,2]{label=(\alph*),ref=\thequestionsi(\alph*), leftmargin=*}

\usepackage{hyperref} 

\usepackage[table,xcdraw]{xcolor}
\usepackage{multirow}
\usepackage{comment}
\usepackage{graphicx}

\usepackage{capt-of}
\usepackage{tabu}
\usepackage{scalerel}
\usepackage{tikz}

\usepackage{blindtext}
\usepackage{geometry}
\title{Explainability in Process Outcome Prediction: Guidelines to Obtain Interpretable and Faithful Models}

\author{
  Alexander Stevens, Johannes De Smedt \\
  Research Centre for Information Systems Engineering (LIRIS) \\
  KU Leuven \\
  Leuven\\
  \texttt{\{alexander.stevens, johannes.desmedt\}@kuleuven.be} \\
}

\begin{document}
\maketitle

\begin{abstract}
Predicting the outcome of various sorts of processes has seen a strong uptake in recent years due to the advances in machine and deep learning. Although a recent shift has been made in the field of process outcome prediction to use models from the explainable artificial intelligence field, the evaluation still occurs mainly through predictive performance-based metrics, thus not accounting for the explainability, actionability, and implications of the results of the models. This paper addresses explainability through the properties \emph{interpretability} and \emph{faithfulness} in the field of process outcome prediction. We introduce metrics to analyse these properties along the main dimensions of (business) process data: the event, case, and control flow attributes. This allows comparing inherently created explanations with post-hoc explanations techniques, for which we benchmark seven classifiers on thirteen real-life events logs covering a range of transparent and non-transparent machine learning and deep learning models complemented with post-hoc (model-agnostic or model-dependent) explainability techniques. Next, this paper contributes a set of guidelines named X-MOP which allows selecting the appropriate model by providing insight into how the varying preprocessing, model complexity, and explainability techniques typical in process outcome prediction influence the explainability of the model. 
\end{abstract}

\keywords{Explainable Artificial Intelligence \and Process Outcome Prediction, Interpretability \and Faithfulness \and Deep Learning \and Machine Learning}

\section{Introduction}\label{sec:introduction}
Both in operations research (OR) and business process management (BPM), prevalent topics include the modelling of processes in order to identify possible problems such as bottlenecks caused by a mismanagement or lack of resources \cite{DBLP:journals/mansci/DawandeFJ21} with the goal to find root causes in the process flow \cite{land2021inventory}. Over the past two decades, the BPM domain has seen a strong uptake of data-driven process analysis, coined under the term process mining, which uses process data generated by executed processes for cases within an information system \cite{DBLP:books/sp/Aalst16}. This follows a similar trend in OR, where research shifted because of the access to large databases on (operational) transactions and a lack of back testing~\cite{DBLP:journals/mansci/Graves21}. The focus of this study lies with predictive process monitoring~\cite{DBLP:journals/tkdd/TeinemaaDRM19}, the umbrella term geared towards process mining for predictive activities. It allows identifying process-related trends regarding particular outcomes (e.g., will customers be awarded credit?), impeding bottlenecks (e.g., how long will it take to process my credit application?), and whether particular activities will occur in the future (e.g., will a credit check be necessary for this application?). When the concrete objective is to predict the outcome of an incoming, incomplete case, the field of study is referred to as Process Outcome Prediction (POP). The process data used in this research field is also referred to as \emph{event logs}, as the occurrence of a single activity in a process (case) is referred to as `event'. Moreover, an event log consists of traces, each a sequence of events produced in the context of one case. In the situation of a loan application process, each event records the occurrence of an activity (with the activity label being the \emph{control flow attribute}) up until the loan request is either accepted or rejected. Each activity can have complementary \emph{case attributes} and \emph{event attributes} such as the requested loan amount (case) and the current officer (event) respectively. These case attributes remain unchanged (i.e. \emph{static}) within each case and hence in the whole trace, while the event attributes have varying values (i.e. \emph{dynamic}) for every event. This means that an event possibly has three different attribute types: event, case, and control flow attributes. 
In recent years, a wide array of traditional machine learning models have been used for process outcome prediction~\cite{DBLP:journals/tkdd/TeinemaaDRM19}. An often anticipated trend is the introduction of computationally complex models in order to improve the predictive performance~\cite{kratsch2020machine}. The use of such complex \emph{black box} models comes at the cost of obfuscating their inner workings and therefore being unable to provide insights into why a certain business process prediction was made. The field of eXplainable Artificial Intelligence (XAI) focuses on gaining insights into \emph{how} and \emph{why} certain predictions were made, while trying to maintain the predictive performance of these highly performant models~\cite{DBLP:journals/inffus/ArrietaRSBTBGGM20}. However, many XAI proponents state that trying to explain these black box models comes with a loss of \emph{faithfulness} (or fidelity) due to the fact that the relationship between input and output is only an approximation~\cite{rudin2019stop,DBLP:journals/jbi/MarkusKR21,zhou2021evaluating}. Nonetheless, the use of increasingly complex models has been widely adopted in high stake decision-making processes throughout society~\cite{rudin2018optimized}.


This paper identifies and addresses two main issues that arise at the intersection of POP and XAI. 
First, the existing POP research is constrained to the predictive performance while neglecting the interpretation, actionability, and implications of the results. Many of the existing explainability metrics are either not model-agnostic (i.e. dependent on model parameters), not able to compare transparent versus black box models, or not adapted to a process-based analysis (i.e., do not take into account the different dimensions of process data). 
In addition, recent literature pointed out that the post-hoc explainability techniques should uncover, apart from an interpretable explanation, the \emph{true reasons} for model predictions. Hence, a more comprehensive study about the faithfulness and interpretability of XAI techniques in POP is missing, which impedes the selection of the most appropriate predictive algorithms and explainability models to achieve an interpretable outcome. We address this issue by introducing model-agnostic explainability metrics based on the properties interpretability and faithfulness, which are able to assess and compare transparent models versus black box models that require post-hoc model-agnostic (or model-dependent) explainability methods. 
These metrics are not yet available in the field of XAI, and are analysed along the event, case, and control flow perspective, making them suitable for evaluation purposes in the field of POP. We perform a benchmarking study with five machine learning (ML) and two deep learning (DL) algorithms with different explainability approaches. In this sense, this paper is complementary with the study of~\cite{DBLP:journals/tkdd/TeinemaaDRM19} and~\cite{kratsch2020machine}, as it compares the experimental set-up with both deep learning and newly-introduced interpretable models, i.e., the Logit Leaf Model (LLM) \cite{de2018new} and the Generalized Linear Rule Models (GLRM) \cite{wei2019generalized}. 
We also provide generic, model-agnostic implementations for the XAI metrics which are also compatible with the latter models.
Hence, this study adds the XAI aspect and reconsiders what models excel in terms of predictive accuracy, interpretability, and faithfulness.
Second, the main focus of this field is to verify its conformance with respect to the business requirements and goals~\cite{nunes2017systematic,DBLP:books/sp/DumasRMR18}, but there remains a need for a set of guidelines towards obtaining faithful and interpretable explanations in the context of business process monitoring. We provide a framework to obtain accurate and eXplainable Methods for process Outcome Prediction (X-MOP). This framework of guidelines is based on the evaluation of a wide benchmark of preprocessing, traditional machine learning, deep learning and explainability approaches. 

This work extends the initial work of~\cite{DBLP:conf/icpm/StevensSP21}, focused on comparing inherently created explanations with post-hoc explanations in the context of POP. This paper is a considerable extension with new event logs, new algorithms and new XAI metrics. Next, we further adapt the original metrics along the control, case, and event perspective, which is typical for a process-based analysis. Finally, this papers has bundled the insights obtained from the wide benchmark into a framework of guidelines to obtain accurate and explainable models for POP. The rest of the paper is organized as follows. First, a review of the literature regarding explainability in predictive process monitoring is given in Section~\ref{sec:related work}, together with the motivation for this line of research. This is followed by preliminaries defined in Section~\ref{sec:preliminary}. Next, a definition for explainability is given, together with the introduced metrics in Section~\ref{sec:explainabilityPOP}. The benchmark study and implementation details can be found in Section~\ref{sec:benchmarkstudy}, after we clarify the research questions in Subsection~\ref{sec:RQ}. In Section~\ref{sec:evaluation}, the insights obtained from the research questions are incorporated into the framework of guidelines named X-MOP. Finally, the results and obtained insights are concluded in Section~\ref{sec:conclusion}.

\section{Related Work and Motivation}\label{sec:related work}

Predictive process monitoring is concerned with providing insights about the business processes of modern organizations. Most predictive efforts are primarily driven by using classical machine learning (ML) approaches such as logistic regression (LR)~\cite{DBLP:journals/tkdd/TeinemaaDRM19}, ensemble methods such as XGBoost (XGB) and Random Forest (RF)~\cite{leontjeva2016complex, senderovich2017intra}, with recent works showing interest in applying deep learning models~\cite{wang2019outcome,weytjens2020process}. In~\cite{wang2019outcome}, they propose to use attention-based bidirectional Long Short-Term Memory (LSTM) neural networks for process outcome prediction. Next, \cite{weytjens2020process} compare LSTMs with Convolutional Neural Network (CNN), stating that the latter models work better. The study of~\cite{DBLP:journals/tkdd/TeinemaaDRM19} provides an extensive benchmark that comprises different event logs and classic approaches used for process outcome prediction, with the study of~\cite{kratsch2020machine} describing which event log properties facilitate the use of deep learning (DL) methods. 

Nonetheless, the lack of transparency of these sophisticated models prohibits the ability to understand the rationale of the decision-making process. Over the last two years, the topic of explainability has gained traction in predictive process monitoring. The different works are often divided into two different trends based on how they deal with the explainability-performance trade-off~\cite{DBLP:journals/inffus/ArrietaRSBTBGGM20}. The first trend generates \emph{post-hoc} explanations for black box models. Several papers have already suggested explainability techniques on top of machine learning models, such as Local Interpretable Model-Agnostic Explanations (LIME) SHapley or Additive exPlanations (SHAP) values~\cite{el2022xai,DBLP:conf/icsoc/SindhgattaOM20}, with similar developments in a deep learning context. As an example, \cite{ExplainablePredictiveProcessMonitoring} visualizes the influence of certain attributes in the different steps of the process in a LSTM model with the use of SHAP values. Next, \cite{mehdiyev2021explainable} focuses on creating local post-hoc explanations with the use of a surrogate decision tree. In~\cite{InterpretablePredictiveModels}, Bidirectional LSTMs are used, where the hidden states of the time steps of both RNNs are concatenated. After this, a context vector is learnt that takes the different time steps into account.~\cite{harl2020explainable} visualizes the impact of the activities on the predictions with the use of gated graph neural networks. 
The second trend introduces interpretable models instead of trying to \emph{break open} these black box models, stating that there are alternative models that yield better explainability-performance trade-offs. In~\cite{pasquadibisceglie2021fox}, a set of fuzzy rules are learnt from neural networks. Here, the relationship between inputs and output is determined by a set of IF-THEN rules. Nonetheless, this approach requires domain knowledge in order to bin the attributes into different interpretable terms. Next, a Bayesian network is used in~\cite{pauwels2020bayesian} for next event and suffix prediction. Even though the causal relationships are inferred from historical data, it relies on domain knowledge assumptions. Finally, \cite{DBLP:conf/icpm/StevensSP21} draws two advanced logistic regression models from the XAI literature and adapts them to POP. These models, the Logit Leaf Model (LLM) and the Generalized Logistic Rule Model (GLRM), are originally introduced by \cite{de2018new} and \cite{wei2019generalized} respectively. 

In the field of XAI, a wide range of works have already evaluated these different models with predictive performance-based metrics or with metrics that assess the quality of the explanation methods. As an example, \cite{DBLP:conf/flairs/IslamEG20} created a metric for explainability based on \emph{human-friendly properties}, while~\cite{DBLP:conf/pkdd/MolnarCB19} introduced a metric based on three different properties that define \emph{model complexity}. Nonetheless, these metrics do not take into account the different attributes that are typical of a process-based analysis. The benchmark study of~\cite{el2022xai} compares different explainability models in the field of predictive process monitoring, but the properties evaluated, \emph{stability} and \emph{duration of execution}, are not directly related to explainability and/or take into account the dimensions of process data. Furthermore, none of the above metrics evaluate the faithfulness of the explainability model. However, work from related fields show that there are substantial problems with the faithfulness of post-hoc explainability methods. First, the work of~\cite{jain2019attention} states that the learned attention weights are uncorrelated with the gradient-based attribute importance, even though they mimic the predictions rather accurately. In addition, they state that a Leave-One-Out (LOO) attribute importance ranking correlates better with the gradient-based attribute importance ranking. Similarly, the findings of~\cite{serrano-smith-2019-attention} show that the importance ranking of the attributes made by attention scores is not faithful to the model decisions (i.e., what the model perceived as important). Second, \cite {ma2020predictive} demonstrated that there is a non-monotonic relationship between the SHAP values and the predictive performance. Third, \cite{velmurugan2021evaluating} introduced metrics to the field of POP that assess the relevance of the attribute value range and the relevance of the decision boundary by calculating the change in predictions by permuting the attribute values inside and outside the attribute value range, respectively. The results show that both LIME and SHAP report low-to-moderate scores for both the faithfulness metrics. A comprehensive study about the (un)faithfulness of explainability methods in POP is hence missing, due to the fact that the accuracy, by which these post-hoc explanations reflect the behaviour of the predictive model, is often inadequate~\cite{rudin2019stop,DBLP:conf/icpm/StevensSP21}. 

To conclude, there is a need for model-agnostic explainability metrics that are adapted to POP and work for both transparent and non-transparent models. Furthermore, the faithfulness of (post-hoc) explainability methods used for POP needs to be evaluated with the use of metrics that are based on the properties of \emph{explainability}, i.e. \emph{interpretability} and \emph{faithfulness}. These domain-specific metrics can guide practitioners to select the best model for the task at hand~\cite{InterpretableMachineLearning} and can reduce the scope of research for human-based studies by reducing the financial and time costs of such experiments.
There is a notable void in the POP literature in this respect, which this paper will address through both introducing the notions of interpretability, faithfulness, and a POP-specific set of guidelines.

\section{Preliminaries}\label{sec:preliminary}
This section first describes the different nomenclatures inherited from the XAI field, followed by the preliminary steps needed for predictive process monitoring. 

\subsection{The different XAI nomenclatures}\label{nomenclatures}

The \emph{task model} is defined by several studies as the predictive model that generates the predictions~\cite{DBLP:journals/jbi/MarkusKR21,rudin2019stop}. The \emph{explainability model} is the model that generates the explanations for the predictions made by the task model. Recent literature describes \emph{transparency} as the opposite of \emph{blackbox-ness}~\cite{DBLP:journals/inffus/ArrietaRSBTBGGM20}.
In the case of a \emph{transparent} task model (also referred to as an \emph{interpretable} model), the model is able to generate its own explanations, where a \emph{black box} model requires the need of an additional explainability model. This means that a transparent model is technically also its own explainability model, while the explainability model of a black box model can be e.g. a surrogate model, attention layer, or SHAP values. Furthermore, \emph{interpretability}, originally described as \emph{comprehensibility} in~\cite{DBLP:journals/inffus/ArrietaRSBTBGGM20}, is the ability to provide an explanation that consists solely out of single chunks of information, preferably in a human understandable fashion. It is often quantified by the related concept of model complexity~\cite{DBLP:journals/inffus/ArrietaRSBTBGGM20,DBLP:conf/pkdd/MolnarCB19}. Note that an~\emph{interpretable model} is different from the~\emph{interpretability}. E.g., an interpretable model (e.g., a logistic regression model) that creates its own explanations based on more than 500 attributes has a low value for interpretability. In the XAI literature, the explainability-accuracy trade-off compares the model interpretability with the model accuracy, assuming that it is required to strike a balance between either simple models (e.g., linear regression) or models using complex inference structures (e.g., neural networks). By contrast, this paper investigates whether the \emph{interpretability of an explanation} is also in trade-off with the \emph{predictive accuracy}. In this sense, a non-interpretable deep neural network can have higher value for the interpretability of explanations compared to a logistic regression model. This is similar to the insights obtained in~\cite{DBLP:journals/jbi/MarkusKR21}. Next, even though often used interchangeably~\cite{DBLP:journals/inffus/ArrietaRSBTBGGM20}, interpretability and \emph{explainability} differ significantly due to the fact that an interpretable explanation is not always faithful. To emphasize, a simple explanation generated for a rain forecast prediction could be: \textit{`if the grass is green, it will rain'}, which is easy to interpret, but unfaithful. The necessity to distinguish between faithfulness and interpretability has already been pointed out by prior research~\cite{DBLP:journals/jbi/MarkusKR21,zhou2021evaluating}. The \emph{faithfulness} of an explainability model can be considered as the accuracy by which the explainability model accurately mimics the behaviour of the task model (and \emph{not} the predictions of the task model), as similar predictions do not ensure that the behaviour of the task model is correctly mimicked~\cite{rudin2019stop}. 
\subsection{PPM setup}
PPM relies on the use of historic process data recorded in an (event) log, which is a set of traces representing the enactment of a process for a particular case (e.g. a loan application) within an information system~\cite{DBLP:books/sp/Aalst16}. The occurrence of a single activity in a process (case) is referred to as `event'. Such an event may record three different attribute types typically used for a process-based analysis: the \emph{control flow attribute} (the activity), the \emph{event attributes} (i.e.~\emph{dynamic} attributes, which may change from one event to another), and the \emph{case attributes} (i.e. \emph{static} attributes, which do not change throughout the lifetime of a case).

An event is a tuple $e = (c,a,t,(s_{1}\dots s_{m_{s}}), (d_{1}\dots d_{m_{d}}))$, with $e \in \xi$ (the event universe), $c$ the case ID, $a \in A$ the activity, and $t \in \mathbb{R}$ the timestamp. This event records both case attributes $S = ((s_{1}\dots s_{m_{s}}))$ and event attributes $D = (d_{1}\dots d_{m_{d}})$, with $m_{s}$ and $m_{d}$ the number of case and event attributes, respectively. A trace is a sequence of events $\sigma_c = [e_{1} \dots e_{n}]$ generated by executing activities in a process, sorted based on the timestamps of the events, such that $\forall i \in{[1 \dots |\sigma_c|]}, e_{i} \in\xi$ and $\forall i_{1}, i_{2} \in{[1 \dots |\sigma_c|]}$ $e_{i_{1}}.c = e_{i_{2}}.c$, i.e., all the events in a trace belong to the same case. Consequently, an event $e_{i}$ in a trace $\sigma_{j}$ of the event log $L$ is denoted as $e_{i,j} = (c_{j},a_{i,j},t_{i,j},S_{j},D_{i,j})$. The outcome $y$ of a trace in the case of POP is usually a binary attribute~\cite{kim2022encoding} and depends on the needs and objectives of the process owner~\cite{DBLP:journals/tkdd/TeinemaaDRM19}. 

In order to learn, preferably incrementally over time as traces are sequences over time, from the development of traces, (trace) prefixes are often extracted from the completed cases. To this end, a prefix log $\mathbb{L}$ is derived, which is the extracted event log $L$ that contains all the prefixes of each case in the original event log. Next, trace cutting, i.e. limiting the prefix up to a certain number of events, is typically performed for computational reasons. Additionally, trace cutting is also performed when the class label of the case is dependent on the occurrence of an event, otherwise the label of the class becomes known and irreversible~\cite{DBLP:journals/tkdd/TeinemaaDRM19}. More information about the labelling of cases is given in Section \ref{sec:event logs}. 
Next, the use of an encoding mechanism enables the user to work with a varying amount of attributes, since each trace can have a different length. An often used encoding mechanism is the aggregation encoding technique~\cite{DBLP:journals/tkdd/TeinemaaDRM19}. First, the categorical static attributes are one-hot encoded, which means that each static attribute that is categorical, results in a number of transformed attributes based on the set of unique attribute values. These values can be found with the use of $\theta$, also referred to as the uniqueness operator. The numeric static attributes remain unchanged. Second, the timestamp is transformed into three different dynamic numeric attributes: \emph{timesincelastevent}, \emph{timesincecasestart} and \emph{timesincemidnight}. Next, all the dynamic (i.e. event) numeric attributes are replaced by their summary statistics $\emph{min}$, $\emph{max}$, $\emph{mean}$, $\emph{sum}$, and $\emph{std}$. The last transformation step relates to the dynamic categorical attributes and the control flow attribute, where the frequency of occurrence of an attribute value in a prefix is the value for the new attribute. By contrast, the use of the above encoding mechanism in step-based models such as recurrent neural networks becomes superfluous given their sequential setup. To exploit this efficiently, a low-dimensional representation of discrete attributes in the form of embeddings is an often performed encoding technique~\cite{evermann2017predicting}. This mapping transforms a categorical attribute to a vector of continuous numbers, similarly to how one-hot encoding works, although the latter ignores the similarity between the obtained vectors. Finally, the tuple $(x_{j,1},x_{j,2},\dots,x_{j,p})$ consists of the resulting attribute values (of all the attributes) of prefix trace $\sigma_{j},j \in [1\dots l]$ with $l$ the number of prefix traces after prefix extraction. Note that the case ID attribute is removed from all the traces. This means that the total number of resulting attributes is p, and $p= p_{a} + p_{s} +p_{d}$, with $p_{a}$, $p_{s}$ and $p_{d}$ the number of control flow, case, and event attributes after the data transformation steps respectively.
This additionally means that the tuple $(x_{j,1}\dots x_{j,p_{a}})$ denotes the attribute values of prefix trace $\sigma_{j}$ for the attribute type \emph{control flow}, followed by $(x_{j,p_{a}+1}\dots x_{j,p_{a}+p_{s}})$ and $(x_{j,p_{a}+p_{s}+1}\dots x_{j,p_{a}+p_{s}+p_{d}})$ for the attribute type \emph{case} and \emph{event} respectively.

Another data transformation step in the context of PPM is referred to as trace bucketing, where traces are divided into different buckets while creating separate models for each of them. This technique is commonly used to support the discovery of heterogeneous segments in the data \cite{di2016clustering,DBLP:journals/tkdd/TeinemaaDRM19}. Algorithms such as K-Nearest Neighbours~\cite{DBLP:conf/caise/MaggiFDG14} or K-Means clustering measure the (dis)similarity between traces depending on the parameter K. The prefix bucketing technique~\cite{leontjeva2016complex} creates different buckets for the prefixes of different lengths, while the state-based bucketing technique~\cite{DBLP:conf/bpm/LakshmananDKCK10} creates a different model for each different decision point within the process model. Although these bucketing techniques can effectively diminish the runtime performance~\cite{DBLP:journals/tkdd/TeinemaaDRM19}, they do not necessarily result in an intuitive or interpretable outcome. E.g., clustering techniques can base their grouping on a high number of dimensions that are not interpretable. Nonetheless, the use of clusters is deemed a necessary action when there is a notion of trace similarity~\cite{de2016general}. The final transformed event log is split into a train and test event log, where the former is used to create a task model to predict the dependent variable based on independent attributes. Moreover, the prediction for a prefix trace $\sigma_{j}$ is denoted as: $\hat y_i = F(x_{j,1},x_{j,2},\dots,x_{j,p})$.

The final step is to interpret the predictions made by the task model $F$. There already exist many model-agnostic methods that try to explain the behaviour of the task model. Moreover, \emph{attribution-based} methods determine the contribution that each attribute made to the final prediction and use this as an explanation. Techniques such as accumulated local effect plots, surrogate models or permutation feature importance are often used in this context. For example, in \cite{molnar2020interpretable}, they calculate the importance of an attribute by calculating the change in the prediction error of the model $F$ before and after permuting the values of an attribute. An attribute is deemed \emph{important} when permuting the values of this attribute increases the prediction error (in mean squared error (MSE)), and is deemed \emph{unimportant} when the MSE remains unchanged after permutation (as the model did not use this attribute for the prediction). The use of this technique is intuitive, as it clearly shows how much \emph{influence} each attribute has.
\begin{center}
\scalebox{0.8}{
\begin{minipage}{1.2\linewidth}
\begin{algorithm}[H]
\caption{Attribute Importance \label{alg:AttributeImportance}}
\begin{algorithmic}
\Require $X \gets (x_{1,1},x_{1,2},\dots,x_{1,p},x_{2,1},\dots x_{l,p}),z \gets [ \ ], w_{PI} \gets [ \ ]$
\For{$i \in [1\dots p]$}
\Comment{loop over all the attributes}
\State $ X^{copy} \gets Copy(X)$ \Comment{make a copy of the original data}
\State $values\_list \gets [ \ ]$
\Comment{empty list for each attribute}
\State $z \gets \theta(x_{1,i},\dots,x_{l,i})$ \Comment{the unique attribute values of attribute $i$}
\For{$j \in [1 \dots l]$} \Comment{loop over all the (prefix) traces}
\State $value \gets x_{i,j}$ \Comment{the current attribute value of instance $i$}
\State $\displaystyle z^{*}\gets\{x^{*} \in z:x^{*}\notin value\}$ \Comment{remove the current value from the list}
\State $values\_list[j] \gets Random(z^{*})$ \Comment{take a random value and add to the list}
\EndFor
\State $X^{copy}_{i} \gets values\_list$ \Comment{replace attribute $i$ values with permuted values}
\State $\hat y \gets predict(X)$
\Comment{original predictions}
\State $ y^* \gets predict(X^{copy})$ \Comment{predictions after permuting attribute $x_{i}$}
\State $MSE_{y,\hat y} \gets \sqrt{\frac{(y - \hat y)^{2}}{n}}$ \Comment{MSE between y and $\hat y$}
\State $MSE_{y,y^*} \gets \sqrt{\frac{(y-y^*)^{2}}{n}}$ \Comment{MSE between y and $y^*$}
\State $effect \gets MSE_{y,y^*} - MSE_{y,\hat y}$ \Comment{attribute importance calculated by change in MSE}
\State $w_{PI}[i] \gets effect$ \Comment{save the effects of all the attributes}
\EndFor
\Return $w_{PI}$ 
\end{algorithmic}
\end{algorithm}
\end{minipage}%
}
\end{center}

 The use of this permutation-based attribute importance technique is described in Algorithm~\ref{alg:AttributeImportance}. Note that some remarks about the design of this algorithm should be made. First, in the case of ML models, permuting the values for original attributes, that are later transformed through frequency aggregation and/or summary statistics, does not make sense, as, e.g., the mean value of attribute values before and after permutation remains the same. Therefore, the attribute importance is calculated for the attributes after transformation $(x_{1},\dots,x_{p})$. On the other hand, in the case of DL models the attribute importance is calculated by permuting the original attributes. The attribute weights defined by this permutation attribute importance technique $PI$ are defined as: $w_{PI} = w_{PI,1},...,w_{PI,p}$.

Likewise, transparent models are able to produce their own attribute importances. As an example, the coefficients in a logistic regression model indicate the importances of the different attributes on the dependent variable. On the other hand, in case of a black box model, the attribute importances can also be measured with the use of a post-hoc explainability model, such as SHAP or LIME, which approximates the attribute weights of the black box model. The attribute importance calculated by the explainability model $E$ (can be both the transparent model itself, or a post-hoc explainability model) is denoted as: ($w_{E,1}\dots w_{E,p}$). 

\section{Explainability in Process Outcome Prediction}\label{sec:explainabilityPOP}

\begin{figure}[t]
\centering
\includegraphics[width=0.7\textwidth]{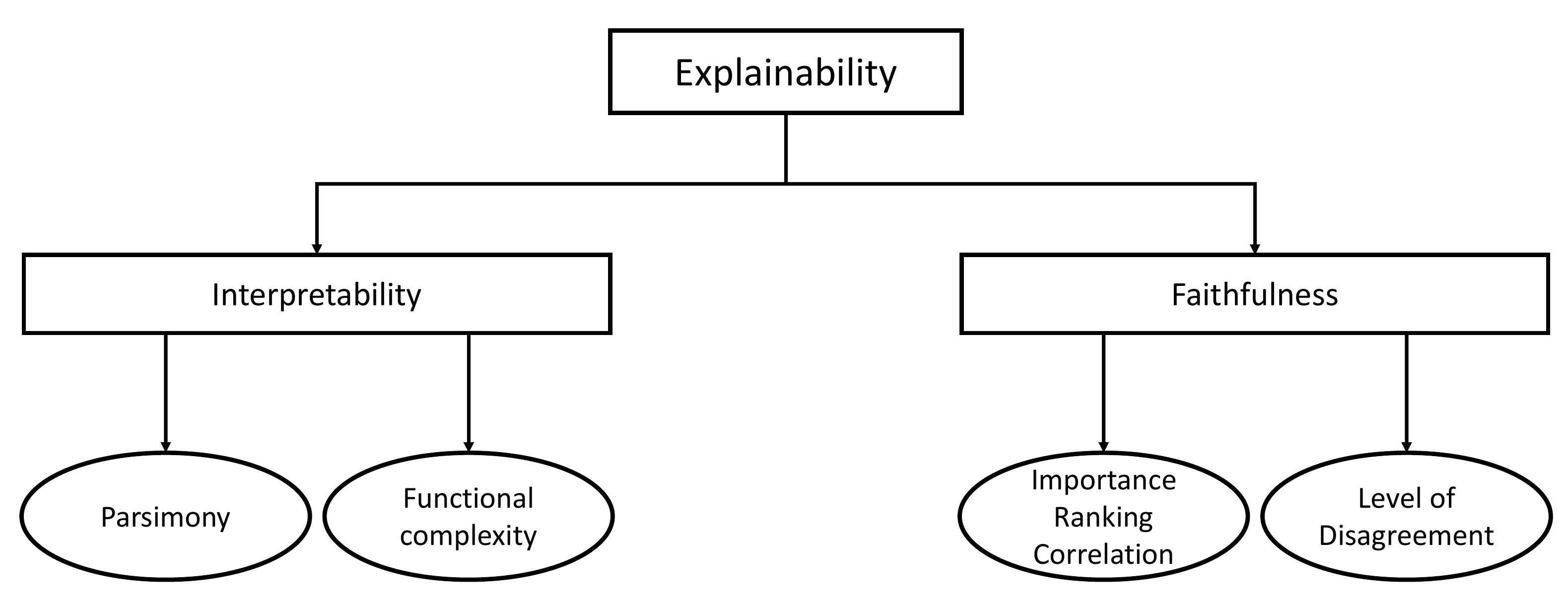}
\caption{Explainability defined through its properties Interpretability and Faithfulness.}
\label{fig:explainabilitydefinition}
\end{figure}

Despite the recent growth of literature in the field of XAI, it still remains unclear what identifies as a suitable explanation and how to evaluate them. One of the reasons is that the quality of an explanation is dependent on many things, such as the use case, the stakeholder, and the explanation method itself~\cite{zhou2021evaluating}. Although \emph{explainability} is a subjective matter, many papers have already stated that there exists properties of explainability (which can be objectively quantified) that a \emph{good} explanation should satisfy. These papers often make the distinction between interpretability and faithfulness, and identify different metrics to quantify these properties~\cite{DBLP:journals/jbi/MarkusKR21,zhou2021evaluating}. Nonetheless, most of the available metrics are not model-agnostic (e.g. dependent on model parameters), or do not work for both transparent and non-transparent models. Furthermore, it is necessary to divide the attributes into the different attribute types, allowing the metrics to take into account the process-based perspectives. 
In this paper, the separation of explainability into interpretability and faithfulness is used to obtain model-agnostic metrics suitable for a process-based analysis. These metrics consequently allow defining and assessing explainability in the context of POP, by being able to quantitatively compare transparent versus non-transparent models extended with explainability methods. An overview of the different properties and metrics introduced by this paper is given in Figure~\ref{fig:explainabilitydefinition}. For more information about the different XAI definitions, properties and/or metrics, we refer to~\cite{DBLP:journals/inffus/ArrietaRSBTBGGM20}, \cite{DBLP:journals/jbi/MarkusKR21} and~\cite{zhou2021evaluating}.

\subsection{Parsimony}\label{sec:parsimony}

Parsimony ($C$) is a property of interpretability that represents the complexity of a model, and is an often used metric for linear regression models. This can be seen as the number of attributes with non-zero weights defined by the explainability model $E$. These are non-zero coefficient weights in a linear regression model, or for a black box model, the non-zero weights provided by the post-hoc explainability model.

First, we reconsider the definition of parsimony as a property for the \emph{interpretability of an explanation} rather than a property for \emph{model interpretability}. Next, we adapt this metric to take into account the different perspectives of a process-based analysis by determining the parsimony for each attribute type separately. The parsimony of the total model $C_{F}$ is equal to the sum of the values for parsimony of the different attribute types. Moreover, a parsimonious (i.e. simple) model corresponds to a low value $C_{F} = C_{control} + C_{event}$ + $C_{case}$.

Assume an attribute $x_{i}$ with $w_{E,j}$ the weight for that attribute $i \in{\{1\dots p\}}$ defined by the explainability model $E$. Then, the parsimony for the control flow attribute type is calculated as follows:
\begin{equation}
C_{control} 
=\sum_{i=1}^{p_{a}} C(x_{i})
\text{\quad with \quad} 
C(x_{i}) = \begin{dcases*}
1, & if $ w_{E,i} > 0 $,\\
0, & otherwise. 
\end{dcases*}
\end{equation}

\noindent The parsimony for the other attribute types is calculated analogously. 

\subsection{Functional Complexity} 
Functional complexity ($FC$) is a measure for model complexity in line with other FC calculations such as the one described by~\cite{DBLP:conf/pkdd/MolnarCB19} and is made suitable for POP data below as\cite{ghorbani2019interpretation} states that small random perturbations to data such as test images can drastically change the generated explanations, without the predicted label being altered. For this, we adapt the metric further by investigating \textit{how many} altered predictions there would be when permuting the attribute values of an attribute type, and consequently measures how strongly the explanations depend on that attribute type. This permutation-based metric is consequently used to retrieve the dependency of the attribute types (i.e., event/case/control flow) on the explanations. In Algorithm~\ref{alg:FC}, the pseudocode for the proposed FC algorithm is given and how the different permutations are made in order to obtain the FC values for the attribute type $control$. Again, the FC for the other attribute types is calculated analogously. 
 
\begin{center}
\scalebox{0.8}{
\begin{minipage}{1.2\linewidth}
\begin{algorithm}[H]
\caption{Functional Complexity of \emph{Control Flow}}\label{alg:FC}
\begin{algorithmic}[]
\Require $X_{test} \gets (x_{1,1},x_{1,2},\dots,x_{1,p},x_{2,1},\dots x_{l,p}), FC_{control}\gets 0$ 
\State $X^{test,copy} \gets Copy(X^{test})$ \Comment{make a copy of the original test data}
\For{$i \in [1\dots p_{a}]$} \Comment{loop over the control flow attributes}
\State $values\_list \gets [ \ ]$
\Comment{empty list for each attribute}
\State $z \gets \theta(x_{1,i},\dots,x_{l,i})$ \Comment{the unique attribute values of attribute $i$}
\For{$j \in [1 \dots l]$} \Comment{loop over all the (prefix) traces}
\State $value = x_{i,j}$ \Comment{the current attribute value of instance $i$}
\State $\displaystyle z^{*} \gets \{x\in z:x\notin value\}$ \Comment{remove the current value from the list}
\State $values\_list[j] \gets Random(z^{*})$ \Comment{take a random value and add to the list} 
\EndFor
\State $X^{test, copy}_{i} \gets values\_list$ \Comment{replace attribute $i$ values with permuted values}
\EndFor
\State $\hat y_i = F(X_{test})$ \Comment{original predictions}
\State $\hat y^* = F(X^{test, copy})$ \Comment{predictions after permuting all control flow attributes}
\State $FC_{control} \gets \frac{distance(\hat y, y^*)}{n}$ \Comment{FC for attribute type control}\\
\Return $FC_{control}$
\end{algorithmic}
\end{algorithm}
\end{minipage}%
}
\end{center}

The algorithm starts by looping over the attributes of the \emph{control flow} attribute type. For each attribute $i$, a random new value is assigned for each prefix trace $j$. The difference with the attribute importance (Algorithm~\ref{alg:AttributeImportance}) is that all the attributes are permuted simultaneously. Next, the predictions before and after these permutations are calculated. Finally, the Hamming Distance, which calculates the number of bit positions in which the two bits (i.e. predictions for a prefix trace) are different, is used as the measure between the different prediction vectors. A low value for $FC_{t}$ means that the predictions are created seemingly independently of this attribute type, and should therefore not be regarded as an important attribute type when interpreting the explanations.

\subsection{Importance Ranking Correlation (IRC)} ($IRC$) is a measure for the faithfulness of an explainability method and is quantified with the use of the non-parametric Spearman's correlation coefficient. The original metric~\cite{nguyen} is model-dependent on a neural network model. Therefore, this paper has adapted the metric to make it model-agnostic by evaluating how faithful the attribute importance ranking of the explainability model is to the ranking made by the permutation attribute importance (PI). Furthermore, this metric does not make a distinction between the different attribute types, as the focus is on the relative ranking of the attributes in general. For a transparent model such as LR, we quantify the $IRC$ between the ranking of the attributes measured with PI and the ranking made by the explainability model (the LR coefficients). For a post-hoc explainability model, we quantify the IRC between the ranking of the attributes measured with PI and the ranking made by the explainability model (e.g. SHAP values). The $IRC$ is defined as: 
\begin{center}
\begin{minipage}{1\linewidth}
\begin{equation}\label{IRC}
IRC=\rho(w_{PI},w_{E})
\end{equation}
\end{minipage}%
\end{center}
with $w_{PI}=w_{PI,1},\dots,w_{PI,p}$ the attribute weights measured with PI and $w_{E} = w_{E,1},\dots, w_{E,p}$ the weights of the explainability model. This correlation coefficient takes a value between [-1,1] and describes the association of rank. A perfectly faithful model has a correlation coefficient of +1, where a loss in faithfulness corresponds with a value closer to 0. Consequently, a negative value corresponds to a negative rank association between the two attribute importance weights. 

\subsection{Level Of Disagreement (LOD@10)} ($LOD@10$) is a metric of faithfulness defined by~\cite{DBLP:journals/corr/LakkarajuKCL17}, which computes the percentage of similar predictions between attribute importance measured with PI and explainability model. 
In predictive process monitoring, it is useful to know the importance of the attribute types, as the obtained insights are relevant in order to improve the early prediction problem~\cite{DBLP:journals/tkdd/TeinemaaDRM19}. In this paper, the LOD@10 therefore investigates whether the PI and the explainability model focus on the same attribute type. The $LOD@10$, different from IRC, neglects the ranking of the attributes but only looks at the relative frequency of the attribute types in the top ten most important attributes (based on their weights). For this, the metric is quantified with the Euclidean Distance between the relative frequency of top ten attributes of the PI and the explainability model, with $w_{PI}^{10} = argmax_{w_{PI}' \subset w_{PI}, |w_{PI}'| =10} \sum_{w \in w_{PI}'}w$ and $w_{E}^{10} = argmax_{w_{E}' \subset w_{E}, |w_{E}'| =10} \sum_{w \in w_{E}'}w$ indicating the top ten highest weights measured with PI and by the explainability model respectively. Next, the control flow attributes that are in the top ten attributes determined by PI and the explainability model are defined as follows: $w_{PI,control}^{10} = {|\{w_{PI}^{10} \mid w_{PI}=w_{PI,1},\dots,w_{PI,p_{a}}}\}|$ and $w_{E,control}^{10} = {|\{w_{E}^{10} \mid w_{E}=w_{E,1},\dots,w_{E,p_{a}}}\}|$. The calculation is similarly for the other attribute types. As an example, the $LOD@10$ is 1.41 when the relative ranking made by the PI is $(w_{PI,control}^{10},w_{PI,case}^{10},w_{PI,event}^{10}) = (1,2,7)$ and the relative ranking of the explainability model $E$ is$(w_{E,control}^{10},w_{E,casel}^{10},w_{E,event}^{10}) = (2,2,6)$.

\begin{center}
\scalebox{0.9}{
\begin{minipage}{1\linewidth}
\begin{equation}\label{LOD@10}
\begin{split}
LOD@10
& = \sqrt {\left(w_{PI,control}^{10}-w_{E,control}^{10}\right)^2+\left(w_{PI,case}^{10}-w_{E,case}^{10}\right)^2+\left(w_{PI,event}^{10}-w_{E,event}^{10}\right)^2}\\
\end{split}
\end{equation}

\begin{align} 
& \text{with \space \space} \
\begin{dcases*} 
w_{PI,control}^{10}+w_{PI,case}^{10}+w_{PI,event}^{10} = 10\\
w_{E,control}^{10} \ \ + w_{E,case}^{10} \ + w_{E,event}^{10} \ = 10\\ 
\end{dcases*}
\end{align} 

\end{minipage}%
}
\end{center}

This metric is introduced to take into account that the number of attributes used in the task model has a negative influence on the IRC value. Furthermore, a high value for this LOD@10 metric indicates that the explainability model focus on rather different attribute types, which impairs the faithfulness of the explainability model.

\section{Experimental Setup}\label{sec:benchmarkstudy}

In this section, a detailed build-up to the research questions related to the interpretability and faithfulness of POP models is provided, which will be used in a benchmark setting. Next, the different event logs and their specifications are described, followed by a description of the benchmark models and explainability techniques often used for POP purposes. Finally, the hyperoptimization settings and implementation details of the different setups are given.

\subsection{Research Questions}\label{sec:RQ}

This paper aims at investigating the influence of the most important steps in an POP context on the predictive performance and XAI metrics from Section~\ref{sec:explainabilityPOP}, in order to establish a set of guidelines to obtain accurate and explainable POP solutions. For this, an experimental pipeline similar to the benchmark studies of~\cite{DBLP:journals/tkdd/TeinemaaDRM19} and~\cite{kratsch2020machine}, extended with an XAI dimension, is used. 
To this purpose, the following research questions are investigated:
\begin{questions}
 \item \textbf{How do the different POP methods compare in terms of interpretability versus predictive performance?}
 \item \textbf{What do the different XAI metrics tell about the inherent architecture of the methods?}
 \item \textbf{How should the faithfulness of explanations be evaluated when compared with interpretability and predictive performance?}
\end{questions}
The first research question (RQ1) investigates the trade-off between interpretability (measured with \emph{parsimony}) and predictive performance. This additionally allows assessing whether transparent models typically underperform compared to the black box models in the case of sequential and high-dimensional data~\cite{kratsch2020machine}.

The second research question (RQ2) investigates how the different attribute types (i.e. \emph{event}, \emph{case} and \emph{control flow}) relate to the different XAI metrics and the different model architecture of the POP methods. We assess whether deep learning models focus more on the control flow perspective (i.e. higher $C_{control}$) compared to the traditional cross-sectional statistical and machine learning models, as their sequential architecture is more tailored towards modelling time-dependent and sequential data tasks without such aggregation. Next, we compare the interpretable LLM model, introduced to POP in~\cite{DBLP:conf/icpm/StevensSP21}, as an alternative to the bucketing technique (see Section \ref{sec:preliminary}) in conjunction with the LR model based on the performance and four explainability metrics. Finally, the parsimony values of the different attribute types are compared with the values for functional complexity. Although both metrics are a measure for model complexity, the parsimony evaluates the importance of the attribute types as perceived by the task model (calculated on the training data), while the functional complexity investigates the importance of the attribute types on the predictions (calculated on the test data). 

The third research question (RQ3) investigates how we should interpret the different faithfulness metrics, and how they relate to the interpretability metrics and the predictive performance. First, we assess whether the predictive outperforming of certain models comes at the expense of the interpretability and/or faithfulness. Next, we investigate whether the explanations generated by post-hoc explainability
techniques (e.g. SHAP values) are less faithful compared to the explanation method that contribute to the predictions of the black box model (e.g. attention values). Moreover, we implicitly assume that the faithfulness of the latter should be higher compared to the SHAP values. 

\subsection{Event logs}\label{sec:event logs}
\begin{table}[ht]
\caption{The Different Specifications of the Event Logs.}
\label{tab:eventlogspecs}
\resizebox{\textwidth}{!}{%
\begin{tabular}{|c|c|c|c|c|c|c|c|c|c|c|c|c|c|}
\hline
\rowcolor[HTML]{EFEFEF} 
Event Log & \textit{Traces} & \textit{Events} & \textit{Med.} & \textit{Max.} & \textit{Prefix} & \textit{Var.} & \textit{$\frac{Act}{Trace}$} & \textit{Stat. cat.} & \textit{Dyn. cat.} & \textit{$\frac{Var.}{Trace}$} & \textit{$\frac{Events}{Trace}$} & \textit{$\frac{Events}{Act}$} & \textit{$\frac{Dyn.}{Stat.}$} \\ \hline
\cellcolor[HTML]{C0C0C0}\textbf{BPIC2011(1)} & 1140 & 67480 & 25 & 1814 & 36 & 815 & 193 & 961 & 290 & 0.71 & 59 & 3 & 0.3 \\ \hline
\cellcolor[HTML]{C0C0C0}\textbf{BPIC2011(2)} & 1140 & 149730 & 54.5 & 1814 & 40 & 977 & 251 & 994 & 370 & 0.86 & 131 & 5 & 0.37 \\ \hline
\cellcolor[HTML]{C0C0C0}\textbf{BPIC2011(3)} & 1121 & 70546 & 21 & 1368 & 31 & 793 & 190 & 886 & 283 & 0.71 & 63 & 3 & 0.32 \\ \hline
\cellcolor[HTML]{C0C0C0}\textbf{BPIC2011(4)} & 1140 & 93065 & 44 & 1432 & 40 & 977 & 231 & 993 & 338 & 0.86 & 82 & 3 & 0.34 \\ \hline
\cellcolor[HTML]{C0C0C0}\textbf{BPIC2015(1)} & 696 & 28775 & 42 & 101 & 40 & 677 & 380 & 19 & 433 & 0.97 & 41 & 1 & 22.79 \\ \hline
\cellcolor[HTML]{C0C0C0}\textbf{BPIC2015(2)} & 753 & 41202 & 55 & 132 & 40 & 752 & 396 & 7 & 429 & 1.00 & 55 & 1 & 61.29 \\ \hline
\cellcolor[HTML]{C0C0C0}\textbf{BPIC2015(3)} & 1328 & 57488 & 42 & 124 & 40 & 1280 & 380 & 18 & 428 & 0.96 & 43 & 1 & 23.78 \\ \hline
\cellcolor[HTML]{C0C0C0}\textbf{BPIC2015(4)} & 577 & 24234 & 42 & 82 & 40 & 576 & 319 & 9 & 347 & 1.00 & 42 & 1 & 38.56 \\ \hline
\cellcolor[HTML]{C0C0C0}\textbf{BPIC2015(5)} & 1051 & 54562 & 50 & 134 & 40 & 1048 & 376 & 8 & 420 & 1.00 & 52 & 1 & 52.5 \\ \hline
\cellcolor[HTML]{C0C0C0}\textbf{SEPSIS(1)} & 782 & 13120 & 14 & 185 & 29 & 684 & 14 & 195 & 38 & 0.87 & 16 & 1 & 0.54 \\ \hline
\cellcolor[HTML]{C0C0C0}\textbf{SEPSIS(2)} & 782 & 10924 & 13 & 60 & 13 & 656 & 15 & 200 & 40 & 0.84 & 14 & 1 & 0.19 \\ \hline
\cellcolor[HTML]{C0C0C0}\textbf{SEPSIS(4)} & 782 & 12463 & 13 & 185 & 22 & 709 & 15 & 200 & 40 & 0.91 & 16 & 2 & 0.2 \\ \hline
\cellcolor[HTML]{C0C0C0}\textbf{Production} & 220 & 2489 & 9 & 78 & 23 & 203 & 26 & 37 & 79 & 0.92 & 11 & 3 & 2.14 \\ \hline
\end{tabular}%
}
\end{table}
This study is based on four different real-life event logs that can be found at the website of 4TU Centre for Research Data\footnote{https://data.4tu.nl/}, and are often used in the field of POP~\cite{DBLP:journals/tkdd/TeinemaaDRM19, kratsch2020machine,harl2020explainable,mehdiyev2021explainable}. These event logs are split with Linear Temporal Logic (LTL) rules as defined in~\cite{DBLP:journals/tkdd/TeinemaaDRM19} to obtain objectives for the process. Moreover, the event log is split based on the labelling functions defined by the four LTL rules, therefore creating four different binary prediction tasks. The event log specifications are defined in Table~\ref{tab:eventlogspecs}. The first two columns indicate the number of traces and events in the event log. The next columns describe the median length, original maximum length and the length after prefix cutting (see~\ref{sec:preliminary}) respectively. The column $Var$ indicates the total number of different activities, with the columns $Stat. cat.$ and $Dyn. cat$ indicating the number of static (i.e. case) and dynamic (i.e. event) categorical attributes, respectively. The remaining columns are intuitive. The first log, BPIC2011, describes the medical history of patients from the Gynaecology department of a Dutch Academic hospital. The applied procedures and treatments of the different cases represent the activities in this event log, with the label being either true or false if the LTL rule is violated or not, respectively. The four different LTL rules to generate the event logs: bpic2011(1), bpic2011(2), bpic2011(3) and bpic2011(4), are described in~\cite{DBLP:journals/tkdd/TeinemaaDRM19}. Next, similar trace prefixing and cutting preprocessing steps are performed as in~\cite{DBLP:journals/tkdd/TeinemaaDRM19}. The second log, BPIC2015, assembles events pertaining to the building permit application process from five Dutch municipalities. A single LTL rule is applied on the event log and split for each of the five municipalities. The LTL rule defines that a certain activity \emph{send confirmation receipt} must always be followed by \emph{retrieve missing data} (and not the other way around), where the latter activity has to always be present in the trace if the former was. No trace cutting was performed on this event log. Next, the sepsis cases event log contains the discharge information of patients with symptoms of sepsis in a Dutch hospital, starting from the admission in the emergency room until the discharge of the patient. Here, the labelling is performed based on the discharge of the patient instead of LTL rules~\cite{DBLP:journals/tkdd/TeinemaaDRM19}. Last, the Production log, contains information about the activities in a manufacturing process, together with the workers and/or machines of the production of the items itself. The labelling function is based on whether the number of work orders rejected is larger than zero or not.
\subsection{Benchmark models and explainability models}
The wide variety of classifiers and explainability models are chosen based on their frequent presence in other studies, such as~\cite{DBLP:journals/tkdd/TeinemaaDRM19,kratsch2020machine,ExplainablePredictiveProcessMonitoring,InterpretablePredictiveModels}. The first model is the logistic regression model, an often used interpretable predictive technique to model the probability of a discrete variable. The two advanced logistic regression models, LLM~\cite{de2018new} and GLRM~\cite{wei2019generalized}, are interpretable models that were introduced to the field of POP in~\cite{DBLP:conf/icpm/StevensSP21}. The former clusters the data with a decision tree and builds linear models in the leave nodes. The latter creates binary rules with a generalized logistic rule model. Here, the probability of being classified as `deviant' is calculated similarly to a regular logistic regression model~\cite{DBLP:conf/icpm/StevensSP21}. 

The next two models are ensemble machine learning models, i.e. XGBoost (XGB) and Random Forest (RF), which are not interpretable models, as the inherent complexity is what bestows their predictive abilities. In the XGB model, weak learners are iteratively improved to a final strong learner by incorporating the loss function of the previous weak learner(s). On the other hand, the RF trains a number of decision trees on various subsets of the data. Different to XGB, the voting scheme is based on the majority votes of predictions. By contrast to the transparent models, ensemble methods require an explainability model. For this, the explanations for the ensemble methods are created with SHAP values~\cite{lundberg2017unified} created with Tree SHAP, which are calculations for each instance-attribute combination based on coalitional game theory. Here, a prediction is explained by assuming that each attribute value is a player in a game, where the prediction is the payout. This model-agnostic technique tells how to distribute the payout among the attributes, as the SHAP values are the average marginal contribution of an attribute value across all possible coalitions. 

Next, two different neural network models are used. The first model is the LSTM, often used for POP~\cite{ExplainablePredictiveProcessMonitoring,DBLP:conf/caise/TaxVRD17,kratsch2020machine} with the long-term relations and dependencies encoded in the cell state vectors, designed to solve the vanishing gradient problem. The advantage of LSTM over classical machine learning models lies in the ability to model time-dependent and sequential data tasks, where the categorical values are encoded in embeddings. 
The second model is a Convolutional Neural Network (CNN) which is a deep-forward artificial neural network with information flow starting from the input layers, through the hidden layers, until the output layer (therefore only in one direction). To the best of our knowledge, this was only implemented by~\cite{weytjens2020process} for POP, but is incorporated in this study due to the frequent use in related fields such as next activity prediction~\cite{pasquadibisceglie2019using}.
Similar to the ensemble methods, the internal representation of a deep neural network (LSTM and CNN), does not allow for inherent explanations of predictions. The use of attention layers is a model-specific post-hoc explainability technique in the strict sense of the meaning, as attention is contributing to the prediction but typically calculated afterwards to obtain attribute importance scores~\cite{bahdanau2014neural}. These attention layers calculate non-negative weights (multiplied by their corresponding representations) for each input that together sums to one, and finally sums the resulting vectors into a single fixed-length representation~\cite{serrano-smith-2019-attention}.
\subsection{Preprocessing steps}
First, an equivalent train-test split as in~\cite{DBLP:journals/tkdd/TeinemaaDRM19} is performed. To this end, a temporal train-test split is performed that ensures that the period of training data does not overlap with the period of the test data, while the events of the cases in the train data that did overlap with the test data are cut. Next, the traces are cut with maximum length as defined in~\cite{DBLP:journals/tkdd/TeinemaaDRM19}.

Then, the aggregation encoding is used to encode the data for the machine learning algorithms. Note that the aggregation encoding is unique to processes and is, together with the three perspectives, the primordial reason to investigate a dedicated POP XAI-based approach. On the other hand, the deep learning models are built to work with sequential models and therefore do not need the use of such a sequence encoding mechanism. Moreover, the deep learning models use an embedding mechanism for categorical attributes, which can be seen as one-hot encoding along with dimensionality reduction. The bidirectional LSTM neural network architecture with attention layer for interpretation purposes stems from~\cite{InterpretablePredictiveModels}. Compared to the original set-up, we have optimized the code to work for multiple dynamic attributes, and multiple numerical attributes (by adding them as input layers to the LSTM model). Finally, the predictive function of~\cite{InterpretablePredictiveModels} is transformed into a binary outcome-oriented prediction by stripping off the final layer and inserting a sigmoid output layer instead. 
In order to compare, ceteris paribus, the overall performance of the LSTM with the CNN, we ensure that both models have a similar set-up. Therefore, the CNN model starts from the same architecture as the LSTM, and the Bidirectional LSTMs are replaced with 1D convolutional layers. Different to the LSTM model, the attention is calculated directly after the input layers and embeddings (similar to~\cite{yin2016abcnn}) and the kernel size is set to be the same as the length of the sequences and the filter as the length of the concatenated input (to ensure that we can extract attention values)~\cite{yin2016abcnn}. Additionally, an extra dense layer with Rectified Linear Unit (ReLU) activation was added before the final dense layer, in order to ensure that the output is correctly linked back to the inputs.

The event logs are filtered out for which the average obtained AUC over all the classifiers was lower than 50 (i.e. \emph{sepsis cases(2)}, as no analysis should be performed where random luck has a better ability compared to the classifiers in order to distinguish between the classes). For the XAI evaluation, we additionally filtered out the event logs that had an average AUC below 75 (sepsis cases(3) and production). This was to overcome the issue that explainability techniques can only perform well when the original task model is performant enough.

\subsection{hyper optimization details}
Finally, a 4-fold cross validation was performed with the use of~\hyperlink{http://hyperopt.github.io/hyperopt/}{hyperopt} for the machine learning models. The hyper optimization for the deep learning models was done by a 10-fold cross validation, as the neural networks need additional evaluation to overcome the plethora of local minima. For the LR, XGB and RF model, similar hyperparameter settings are taken from~\cite{DBLP:journals/tkdd/TeinemaaDRM19}. For the LLM model, we additionally incorporate the maximum depth of the decision tree and the minimum samples per leaf to ensure that the tree splitting does not allow overfitting. According to~\cite{wang2019outcome}, the optimal dropout rate for the bidirectional LSTM with attention was around 0.2. Therefore, we have set the maximal dropout rate to 0.3 (as a margin of error). More detailed information about design implementations and parameters are provided on GitHub\footnote{https://github.com/AlexanderPaulStevens/Evaluation-Metrics-and-Guidelines-for-Process-Outcome-Prediction} to enhance the reproducibility results.
\section{Experimental Evaluation}\label{sec:evaluation}

First, the research questions are answered in-depth. Next, a case-based evaluation in the context of POP is performed with the use of the event log BPIC2015(1). Finally, this section ends with the framework of guidelines for Explainable AI in POP in order to guide the practitioner to the correct model selection. 

\subsection{Benchmark results}\label{sec:results}


\begin{table}[t]
\centering
\caption{Predictive Performance (AUC) per Event Log and Classifier}
\label{tab:AUCperformance}
\resizebox{0.6\textwidth}{!}{%
\begin{tabular}{|
>{\columncolor[HTML]{C0C0C0}}c |c|c|c|c|c|c|c|}
\hline
\cellcolor[HTML]{EFEFEF}Event Log &
  \cellcolor[HTML]{EFEFEF}LR &
  \cellcolor[HTML]{EFEFEF}LLM &
  \cellcolor[HTML]{EFEFEF}GLRM &
  \cellcolor[HTML]{EFEFEF}XGB &
  \cellcolor[HTML]{EFEFEF}RF &
  \cellcolor[HTML]{EFEFEF}LSTM &
  \cellcolor[HTML]{EFEFEF}CNN \\ \hline
\textbf{BPIC2011(1)} &
  95.88 &
  $\underline{\textbf{97.34}}$ &
  91.98 &
  95.59 &
  93.09 &
  75.5 &
  79.69 \\ \hline
\textbf{BPIC2011(2)} &
  96.56 &
  98.18 &
  97.4 &
  98.21 &
  $\underline{\textbf{98.49}}$ &
  86.55 &
  80.67 \\ \hline
\textbf{BPIC2011(3)} &
  98.71 &
  98.55 &
  98.01 &
  $\underline{\textbf{98.9}}$ &
  98.87 &
  78.78 &
  85.01 \\ \hline
\textbf{BPIC2011(4)} &
  88.21 &
  $\underline{\textbf{89.75}}$ &
  80.63 &
  86.46 &
  89.04 &
  87.22 &
  85.83 \\ \hline
\textbf{BPIC2015(1)} &
  93.42 &
  $\underline{\textbf{93.67}}$ &
  89.21 &
  87.96 &
  92.16 &
  90.29 &
  90.66 \\ \hline
\textbf{BPIC2015(2)} &
  94.74 &
  94.99 &
  87.18 &
  $\underline{\textbf{95.65}}$ &
  93.89 &
  94.34 &
  92.55 \\ \hline
\textbf{BPIC2015(3)} &
  $\underline{\textbf{96.13}}$ &
  95.92 &
  93.75 &
  93.71 &
  96 &
  94.56 &
  93.72 \\ \hline
\textbf{BPIC2015(4)} &
  $\underline{\textbf{94.79}}$ &
  94.28 &
  91.33 &
  92.39 &
  93.93 &
  90.7 &
  85.99 \\ \hline
\textbf{BPIC2015(5)} &
  93.54 &
  93.55 &
  90.49 &
  93.78 &
  $\underline{\textbf{94.86}}$ &
  91.99 &
  93.5 \\ \hline
\textbf{SEPSIS(1)} &
  54.03 &
  45.67 &
  47.18 &
  39.38 &
  32.14 &
  $\underline{\textbf{56.27}}$ &
  50.45 \\ \hline
\textbf{SEPSIS(2)} &
  $\underline{\textbf{92.34}}$ &
  89.47 &
  73.04 &
  87.45 &
  83.04 &
  84.69 &
  82.13 \\ \hline
\textbf{SEPSIS(4)} &
  $\underline{\textbf{74.39}}$ &
  63.05 &
  64.72 &
  71.63 &
  73.89 &
  66.29 &
  64.84 \\ \hline
\textbf{Production} &
  59 &
  61.33 &
  58.98 &
  $\underline{\textbf{75.91}}$ &
  71.35 &
  60.55 &
  69.46 \\ \hline
\end{tabular}%
}
\end{table}
The first research question (RQ1) investigates how the different POP models compare in terms of interpretability versus predictive performance. 
Table~\ref{tab:AUCperformance} provides us with an overview of the predictive performance results for the classifiers and event logs. The bottom-up ranking of the average AUC of the classifiers is as follows: GLRM, CNN, LSTM, XGB, RF, LLM, LR. Next, the bottom-up ranking based on the model interpretability (see~\cite{DBLP:journals/inffus/ArrietaRSBTBGGM20} for more information) is: LR, LLM, GLRM, RF, XGB, CNN and LSTM. This means that there is no trade-off between model interpretability and predictive performance, as simple models such as LR perform better than complex black-box structures such as LSTMs. Furthermore, the ML models (86.08 AUC) perform better than the DL models (81.23 AUC) on average. A possible explanation why the DL models underperform compared to the ML models is due to the trace cutting of the prefixes (see Section~\ref{sec:preliminary}). Without trace cutting (i.e.retaining very long prefixes), the DL models could potentially obtain equal (or higher) AUC values than the ML models, as the former are better capable of handling long-term dependencies. Nonetheless, as the field of POP is also labelled as early sequence prediction~\cite{DBLP:journals/tkdd/TeinemaaDRM19}, the use of trace cutting is justified (as we want the prediction as soon as possible). Another possible explanation is that the DL models are disadvantaged, due to the lack of aggregation in their encoding (which seems to be an encoding mechanism that works well, see~\cite{DBLP:journals/tkdd/TeinemaaDRM19})

\begin{figure}[ht]
\centerline{\includegraphics[width=0.80\textwidth]{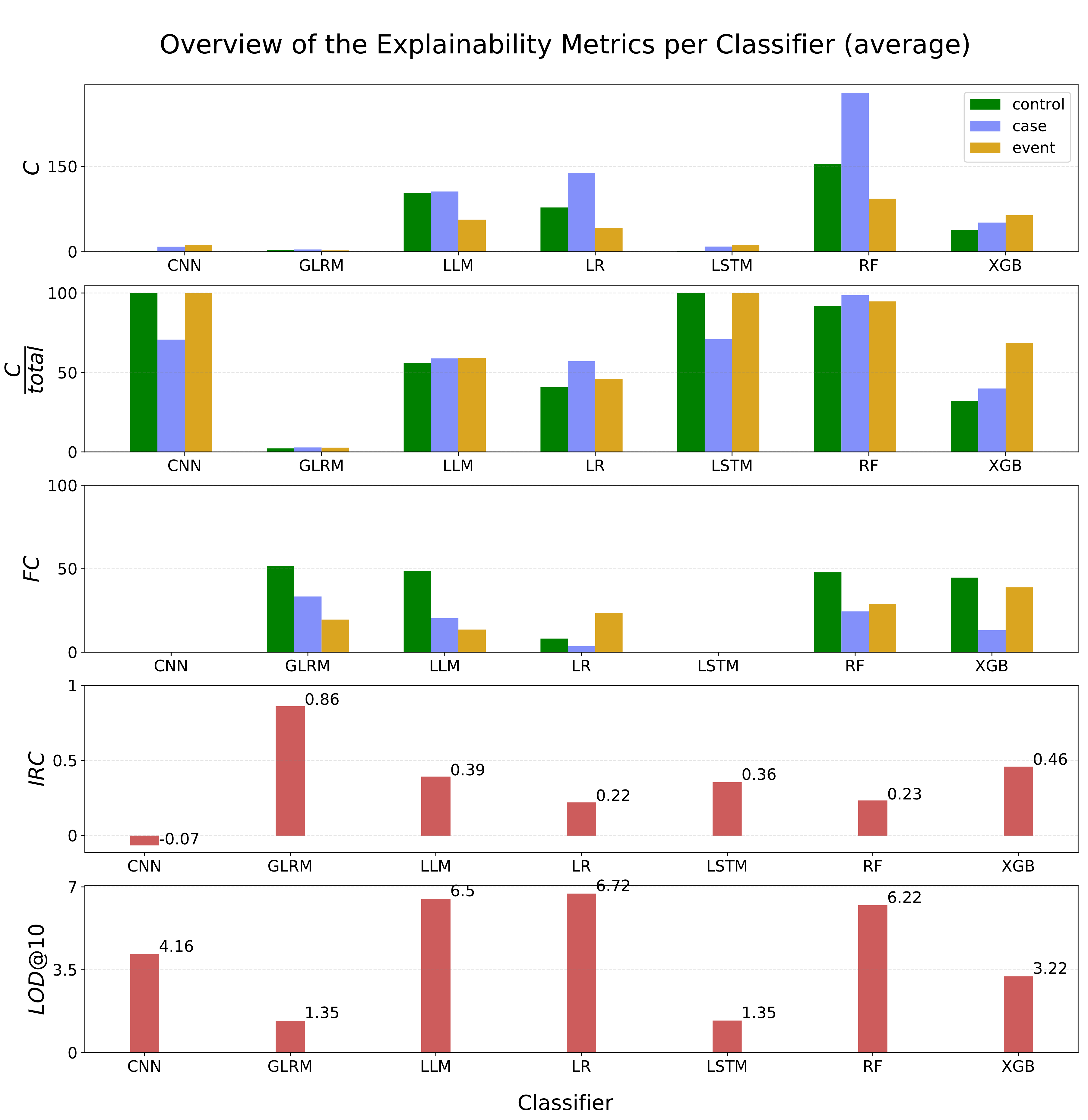}}
\caption{XAI metrics per classifier (averaged over the event logs). The top subplot describes the parsimony of each attribute type (\emph{control},\emph{case} and \emph{event}) individually. The second subplot shows the fraction of the parsimony per attribute type and the total attributes of that attribute type. The third subplot shows the Functional Complexity per attribute type. The fourth and fifth subplot describe the Importance Ranking Correlation ($IRC$) and the Level of Disagreement ($LOD@10$) respectively.}
\label{fig:XAImetrics}
\end{figure}

In~\cite{kratsch2020machine} it is stated that DL models generally outperform classical ML approaches when it comes to process outcome prediction, especially for event logs that have many different activities per trace (i.e. a high value for $\frac{Var.}{Trace}$) and a high ratio of event versus case attributes (i.e. a high value for $\frac{Dyn.}{Stat.}$). When comparing Table~\ref{tab:eventlogspecs} and Table~\ref{tab:AUCperformance}, we see that the DL models obtain accurate results for the BPIC2015 event logs (which have high values for $\frac{Var.}{Trace}$ and $\frac{Dyn.}{Stat.}$), while they perform poorly (compared to the ML models) for the BPIC2011 logs.  Although \cite{kratsch2020machine} used different event logs, it was found that the results apply for logs outside their study due to the purposive sampling of logs and techniques. The only difference is therefore in the sequence encoding mechanism (one-hot encoding instead of aggregation encoding), meaning that the ML model perform better when the dynamic behaviour of dynamic attributes is made static through techniques like frequency aggregation and summary statistics. Furthermore, in line with the results from~\cite{weytjens2020process} and \cite{pasquadibisceglie2019using}, where it was shown that CNN models perform better than LSTM models, we see that CNNs perform better than LSTMs in 5 of 13 event logs. Next, the two transparent models LR and LLM obtain the highest AUC (on average). This means that transparent models are able to attain the performance level of the black box models in the case of sequential and high-dimensional data. Interestingly, the LR (the best overall AUC performance in 4 out of 13 event logs) is ranked higher compared to the LLM model, with the latter model performing better than the LR model in 7 out of 13 event logs. This is due to two reasons. First, the LLM model performs poorly for the sepsis logs. When the sepsis logs are not taking into account (and only the other 10 event logs), then the LLM is ranked higher than the LR. Second, the LLM model is implemented as such that at least one split is \emph{enforced} by the decision tree. This enforced split causes that at least two LR models are created (both on a subset of the data), which might explain the negative effect on the performance of the LLM compared to LR. From the first subplot of Figure~\ref{fig:XAImetrics}, it is clear to see that the CNN, GLRM and LSTM model have significantly lower (better) values for parsimony compared to the other models, with the XGB model as the next in line. Coincidentally, these are the four models with the lowest predictive performance on average. This means that, when only taking into account parsimony as the metric for \emph{interpretability}, the interpretability-predictive performance trade-off holds in the context of POP. In contrast, although the RF model has higher values for $C$ for all the attribute types, it is ranked lower than LR based on predictive performance (AUC).

The second research question (RQ2) investigates what the different XAI metrics tell about the inherent architecture of the models. To start, we can see in the second subplot of Figure \ref{fig:XAImetrics} that only the CNN and LSTM models use 100\% of the event and control attributes. Two interesting remarks can be made for this observation. First, the DL models extract more information from the sequential, control perspective compared to the traditional cross-sectional statistical and machine learning models, as their sequential architecture is more tailored towards modelling time-dependent and sequential data tasks without such aggregation. Second, both the LSTM and CNN also use all the event attributes, which means that these models have greater preference for attributes characterized by their dynamic nature. In contrast, the permutation-based $FC_{event}$ and $FC_{control}$ of the DL models is extremely low. This means that, although the event and control attributes are deemed important by the model (high $\frac{C_{event}}{total_{event}}$ and $\frac{C_{control}}{total_{control}}$), the $FC$ value state otherwise. This casts considerable doubt on the interpretation of the generated explanations, as permuting all the attribute values of the control and event attributes that the model uses has no influence on the prediction made (i.e. the FC remains unchanged). Additionally, interesting insights can be made about whether the relative ranking of parsimony values are similar (or not) to the ranking of the functional complexity values. First, the random forest model seems to use a lot of case attributes (high $C_{case}$), but permuting the values of these attributes does not have a high influence on the predictions (relatively low $FC_{case}$). This shows that the relative ranking of the $FC$ values is not always similar to the relative ranking of the parsimony $C_{F}$ values. This would be a considerable problem in cases where the parsimony indicates that a particular attribute type is relatively important (i.e. $C_{i}=0$), but permuting these values has a very high influence on the predictions (i.e. $FC_{i}>>>0$, meaning that they are still somewhat important). On the other hand, we see in Figure~\ref{fig:XAImetrics} that the FC values of the different attribute types of the GLRM model are rather high, while the parsimony values are rather low. This is desirable, as this means that the model uses few, but influential attributes. 

The third research question (RQ3) investigates how we should evaluate the faithfulness of the explainability models when compared with the interpretability metrics and predictive performance. It is clear to see that the models with the highest average AUC value have the highest value for $LOD@10$, i.e. the LR, LLM, and RF model. We additionally see a relatively low value for $IRC$ for these models. Additionally, we can also see that the LR (the model with the highest ranking based on AUC), is the least faithful based on the metric $IRC$ (apart from the CNN model) and the metric $LOD@10$, whereas the model with the lowest ranking (GLRM) is the most faithful based on these metrics. In addition, we can also see that RF models, which are higher ranked than LSTM model based on AUC, seems to have lower faithfulness values compared to LSTMs. This means that there seems to be a faithfulness-predictive performance trade-off as well. Next, in Section~\ref{sec:preliminary}, it was made clear that the explanation \textit{`if the grass is green, it will rain'} is easy to interpret but unfaithful to the underlying task model behaviour. One could argue that the exploded attribute space of the ML model is detrimental for faithfulness of the model, but the results indicate otherwise. Interestingly, the value for the Importance Ranking Correlation $IRC$ for the XGB(RF) is better in 9(10) out of 13 event logs compared to LSTM, and 12(3) out of 13 compared to the CNN. For the LOD@10, the LSTM only has 1(2) event log where the value is lower when compared with RF(XGB). On the other hand, the CNN model has a lower LOD value in 10(4) event logs in comparison with RF(XGB).  In addition, based on the $IRC$, the LSTM seems to have more faithful explanations than the CNN in 11 of the 13 event logs, only in 9 out of 13 event logs when taking into account the $LOD@10$. The IRC of the CNN model is very compromised (even reports negative values).
Although the inherent architecture of LSTMs are the most suited for the sequential structure of process data, the results show that the faithfulness of post-hoc explanations (e.g. SHAP values) is \emph{less} compromised in comparison with explanations that contribute to the predictions of the black box model but are calculated afterwards (i.e. attention values), which is rather counterintuitive as it is intuitive that the faithfulness of the latter should be higher compared to the SHAP values. This can be due to the fact that attention values focus on explaining the representation of data inside a network rather than explaining the processing of data~\cite{DBLP:conf/dsaa/GilpinBYBSK18}.

To summarize the obtained insights, the introduction of model-agnostic XAI metrics to the field of POP allow obtaining insights into how different POP methods differ based on the properties interpretability and faithfulness. First, we establish that there is a trade-off between predictive performance and interpretability (based on the metric $C$), while there is no trade-off when considering the traditional predictive performance versus model interpretability trade-off. Next, there is a trade-off between predictive performance and faithfulness (measured with $IRC$ and $LOD$), meaning that more faithful models are often at the expense of less accurate models. Finally, the $FC$ (which is a metric to indicate how much predictions change when the attributes of an attribute type are permuted) of control and event attributes of the DL models are almost zero, which means that we can practically change all the values of all the attributes of a certain attribute type, and the predictions will always remain unchanged (which is not something we desire).

\subsection{Event log Analysis: BPIC2015(1)}
\begin{table}[ht]
\caption{Overview of predictive performance (AUC) and XAI metrics for the event log BPIC2015(1).}
\label{tab:bpic2015}
\resizebox{\textwidth}{!}{%
\begin{tabular}{|
>{\columncolor[HTML]{EFEFEF}}c |l|
>{\columncolor[HTML]{FFFFFF}}l |
>{\columncolor[HTML]{FFFFFF}}l |
>{\columncolor[HTML]{FFFFFF}}l |
>{\columncolor[HTML]{FFFFFF}}l |
>{\columncolor[HTML]{C6EFCE}}l |
>{\columncolor[HTML]{C6EFCE}}l |
>{\columncolor[HTML]{C6EFCE}}l |
>{\columncolor[HTML]{FFFFFF}}l |
>{\columncolor[HTML]{FFFFFF}}l |
>{\columncolor[HTML]{FFFFFF}}l |
>{\columncolor[HTML]{FFFFFF}}l |
>{\columncolor[HTML]{FFFFFF}}l |
>{\columncolor[HTML]{FFFFFF}}l |
>{\columncolor[HTML]{FFFFFF}}l |l|l|}
\hline
\cellcolor[HTML]{C0C0C0}\textbf{BPIC2015(3)} & \multicolumn{1}{c|}{\cellcolor[HTML]{EFEFEF}{\color[HTML]{330001} \textit{AUC}}} & \multicolumn{1}{c|}{\cellcolor[HTML]{EFEFEF}\textit{event}} & \multicolumn{1}{c|}{\cellcolor[HTML]{EFEFEF}\textit{ctrl}} & \multicolumn{1}{c|}{\cellcolor[HTML]{EFEFEF}\textit{case}} & \multicolumn{1}{c|}{\cellcolor[HTML]{EFEFEF}\textit{event}} & \multicolumn{1}{c|}{\cellcolor[HTML]{EFEFEF}\textit{$C_{control}$}} & \multicolumn{1}{c|}{\cellcolor[HTML]{EFEFEF}\textit{$C_{case}$}} & \multicolumn{1}{c|}{\cellcolor[HTML]{EFEFEF}\textit{$C_{event}$}} & \multicolumn{1}{c|}{\cellcolor[HTML]{EFEFEF}\textit{$\dfrac{C_{control}}{total_{control}}$}} & \multicolumn{1}{c|}{\cellcolor[HTML]{EFEFEF}\textit{$\dfrac{C_{case}}{total_{case}}$}} & \multicolumn{1}{c|}{\cellcolor[HTML]{EFEFEF}\textit{$\dfrac{C_{event}}{total_{event}}$}} & \multicolumn{1}{c|}{\cellcolor[HTML]{EFEFEF}\textit{$\dfrac{C_{control}}{total_{control}}$}} & \multicolumn{1}{c|}{\cellcolor[HTML]{EFEFEF}\textit{$FC_{control}$}} & \multicolumn{1}{c|}{\cellcolor[HTML]{EFEFEF}\textit{$FC_{case}$}} & \multicolumn{1}{c|}{\cellcolor[HTML]{EFEFEF}\textit{$FC_{event}$}} & \multicolumn{1}{c|}{\cellcolor[HTML]{EFEFEF}\textit{$IRC$}} & \multicolumn{1}{c|}{\cellcolor[HTML]{EFEFEF}{\color[HTML]{330001} \textit{$LOD_{@10}$}}} \\ \hline
\textit{$XGB$} & \cellcolor[HTML]{FCFCFF}{\color[HTML]{330001} 87.96} & \textit{88} & \textit{266} & \textit{35} & \textit{88} & \cellcolor[HTML]{FFFFFF}50 & \cellcolor[HTML]{FFFFFF}21 & \cellcolor[HTML]{FFFFFF}64 & 18.8 & 60 & 72.73 & 18.8 & 91.78 & 5.7 & 18.27 & \cellcolor[HTML]{FAB7B9}0.45 & \cellcolor[HTML]{FCFCFF}{\color[HTML]{330001} 1.41} \\ \hline
\textit{$GLRM$} & \cellcolor[HTML]{DBEFE3}{\color[HTML]{330001} 89.21} & \textit{88} & \textit{266} & \textit{35} & \textit{88} & 4 & 2 & 0 & 1.5 & 5.71 & 0 & 1.5 & 94.81 & 26.68 & 0 & \cellcolor[HTML]{FCFCFF}0.91 & \cellcolor[HTML]{FCFCFF}{\color[HTML]{330001} 1.41} \\ \hline
\textit{$LSTM$} & \cellcolor[HTML]{BEE3CA}{\color[HTML]{330001} 90.29} & \textit{11} & \textit{1} & \textit{17} & \textit{11} & 1 & 1 & 11 & {\color[HTML]{C00000} $\textbf{100}$} & 5.88 & {\color[HTML]{C00000} $\textbf{100}$} & {\color[HTML]{C00000} $\textbf{100}$} & {\color[HTML]{C00000} $\textbf{0.02}$} & 0.01 & \cellcolor[HTML]{FFFFFF}{\color[HTML]{C00000} $\textbf{0.03}$} & \cellcolor[HTML]{F9AFB1}0.4 & \cellcolor[HTML]{FCFCFF}{\color[HTML]{330001} 1.41} \\ \hline
\textit{$CNN$} & \cellcolor[HTML]{B4DFC1}{\color[HTML]{330001} 90.66} & \textit{11} & \textit{1} & \textit{17} & \textit{11} & 1 & 1 & 11 & {\color[HTML]{C00000} $\textbf{100}$} & 5.88 & {\color[HTML]{C00000} $\textbf{100}$} & {\color[HTML]{C00000} $\textbf{100}$} & {\color[HTML]{C00000} $\textbf{0.03}$} & 0.01 & \cellcolor[HTML]{FFFFFF}{\color[HTML]{C00000} $\textbf{0.03}$} & \cellcolor[HTML]{F8696B}-0.07 & \cellcolor[HTML]{F98A8C}{\color[HTML]{330001} 3.61} \\ \hline
\textit{$RF$} & \cellcolor[HTML]{8CCF9E}{\color[HTML]{330001} 92.16} & \textit{88} & \textit{266} & \textit{35} & \textit{88} & \cellcolor[HTML]{FFFFFF}222 & \cellcolor[HTML]{FFFFFF}34 & \cellcolor[HTML]{FFFFFF}86 & 83.46 & 97.14 & 97.73 & 83.46 & 90.32 & 2.13 & 6.34 & \cellcolor[HTML]{F9A6A8}0.34 & \cellcolor[HTML]{F8696B}{\color[HTML]{330001} 4.24} \\ \hline
\textit{$LR$} & \cellcolor[HTML]{6AC181}{\color[HTML]{330001} 93.42} & \textit{88} & \textit{266} & \textit{35} & \textit{88} & 12 & 3 & 0 & 4.51 & 8.57 & 0 & 4.51 & 95.84 & 1.46 & 0 & \cellcolor[HTML]{FABABC}0.47 & \cellcolor[HTML]{FAB3B5}{\color[HTML]{330001} 2.83} \\ \hline
\textit{$LLM$} & \cellcolor[HTML]{63BE7B}{\color[HTML]{330001} 93.67} & \textit{88} & \textit{266} & \textit{35} & \textit{88} & \cellcolor[HTML]{FFFFFF}149.5 & \cellcolor[HTML]{FFFFFF}24.5 & \cellcolor[HTML]{FFFFFF}49.5 & 56.2 & 70 & 55.68 & 56.2 & 98.75 & 4.37 & 0.8 & \cellcolor[HTML]{F9ABAD}0.37 & \cellcolor[HTML]{F98386}{\color[HTML]{330001} 3.74} \\ \hline
\end{tabular}%
}
\end{table}
A more in-depth analysis is given for the event log BPIC2015(1) as an exemplary POP exercise. As mentioned in Section~\ref{sec:event logs}, this event log describes the building permit application process of the first municipality. Here, the label is dependent on whether a certain activity \emph{send confirmation receipt} is always be followed by \emph{retrieve missing data} or not. The label \emph{regular} is given when this rule is followed, and \emph{deviant} otherwise. In table~\ref{tab:bpic2015}, the predictive performance and XAI metrics results for the event log BPIC2015(1) is given. 

First, it is clear to see that the GLRM, LSTM, CNN and LR are the most parsimonious models (lowest $C_{F}$, with $C_{F}=C_{control}+C_{case}+C_{event}$), with a maximum of 15 attributes used in the resulting model. A little remark has to be made for the DL models, as both the LSTM and CNN do not require the aggregation sequence encoding in order to work with sequential data (and their dynamic behaviour). This is visible in Table~\ref{tab:bpic2015}, where the \emph{control}, \emph{case} and \emph{event} columns indicate that the ML models (266 control, 35 case and 99 event attributes) have a much bigger attribute space compared to the DL models (1 control, 17 case, 11 event attributes). Another interesting insight is that, although the DL models use 100\% of their control and event attributes, the $FC_{control}$ and $FC_{event}$ are almost zero. This means that, although the model uses all the control and event attributes, they do not seem to have any influence on the predictions made (as the $FC$ investigates the number of prediction changes after permutating an attribute type). Third, the LTL rule used for this event log is based on the presence of control flow attributes. Intuitively, we expect a high $FC_{control}$ (when we assume that the model has learnt that the label is dependent on the presence of control flow attributes, then this is something we desire), which is clearly visible in the ML models. As an example, the $FC_{control}$ value of the GLRM model for BPIC2015(1) is very high (more than 94\% of the predictions change after permuting the control flow attributes), despite only using 4 control attributes. The $FC_{control}$ values for the DL models, on the other hand, are almost zero, meaning that permuting the attribute values of the control flow attributes does not cause the predictions to change (as the $FC$ investigates the number of prediction changes after permutating an attribute type). Fourth, related to the previous insight, the $C_{control}$ of the LR model is zero, meaning that the LR model did not use any control flow attribute. Therefore, it is clear that the LR did not learn that the label of the event log is dependent on the control flow attributes (although having correct predictions in almost 94\% of the cases?). Fifth, it is also possible to compare the interpretability between different classifiers. Moreover, we see an $FC_{control}$ of 91.78\% for the XGB model, with a value $C_{control}$ of 50. For the RF model, we observe an $FC_{control}$ of 90.32\%, with a value $C_{control}$ of 222 control columns. This means that the XGB model uses fewer attributes compared to RF, but the attributes used are more important for the predictions made (as 91\% percent of the predictions change if these attributes are permuted).

From a predictive performance perspective, the LLM, LR or RF models are preferred. Nonetheless, the $IRC$ and $LOD@10$ values show that the faithfulness of the explainability model of the RF model is compromised the most. These values indicate that the explainability model is not able to correctly mimic the model behaviour of the task model (which is estimated through an attribute permutation importance). This can be due to the fact that the RF model uses all of its attributes based on the values for $\frac{C_{event}}{event}$,$\frac{C_{case}}{case}$ and $\frac{C_{control}}{control}$ (indicated with a red colour). Next, although the LLM model obtains the best predictive performance, it is less interpretable than LR (higher $C_{event}$, $C_{case}$ and $C_{control}$). Nonetheless, the LR did not seem to learn that the label is dependent on the control flow attributes. Finally, the GLRM uses only 4 control attributes ($C_{control}$), and two case ($C_{case}$) attributes, which means that the generated rules are roughly based on the presence (or absence) of certain control attributes in combination with case attributes. The model is still able to obtain a high AUC (89.21) value. This means that the use of the GLRM model is advised, as the model has the best values for both IRC and LOD@10 and is highly interpretable while remaining performant in terms of AUC. This exemplary event log illustrates that the trade-off between predictive performance, interpretability and/or faithfulness is a difficult one. By being able to resort to the full framework of metrics, it becomes possible to make an informed decision on how to approach the POP problem.

\textbf{\begin{figure*}[ht]
\centerline{\includegraphics[width=1\textwidth]{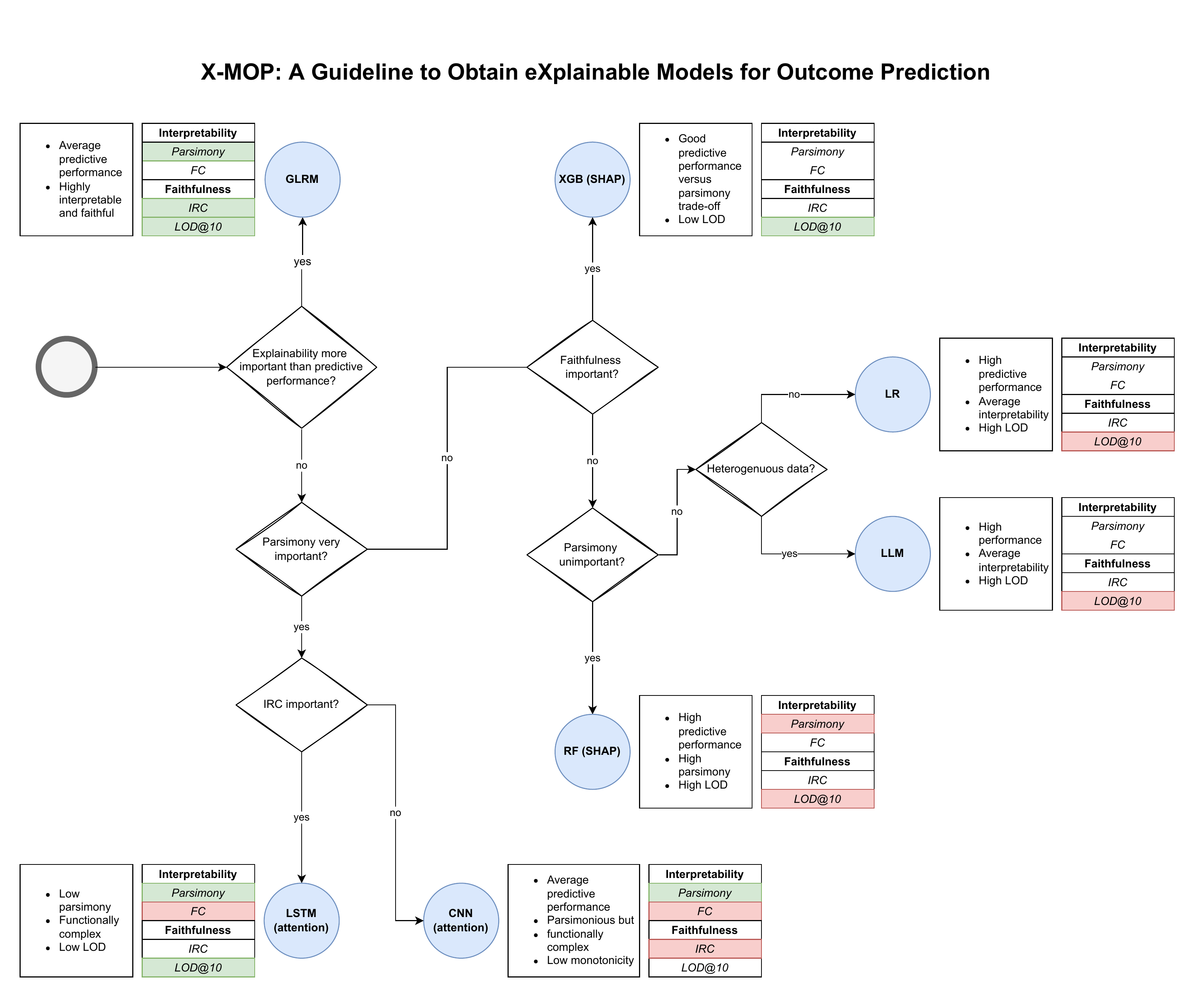}}
\caption{X-MOP: the guidelines for explainable AI purposes in POP}
\label{fig:guidelines}
\end{figure*}}
\subsection{XMOP: Guidelines for XAI in POP}

The insights obtained with the use of the research questions and event log analysis are summarized in an XAI-based guidelines map for POP in Figure~\ref{fig:guidelines}. This figure describes the guidelines for Explainable AI purposes in POP. The white boxes are the questions to guide partitioners and researchers in obtaining POP results according to their preferences in terms of predictive accuracy and explainability. The blue circles are the recommendations for the POP predictive models and explainability models, together with indications whether the advised methods tend to score well (green) or poor (red) for the explainability metrics. Note that these recommendations do not apply to all event logs, but are meant to guide authors to models that comply the best with their requirements (based on the explainability metrics). 

The guideline starts with the start event, whereafter a first question arises: \emph{Is explainability a lot more important than predictive performance?}. If so, the guidelines directly advise using the GLRM model, as this model tends to be parsimonious and faithful, while obtaining an average high AUC. Next, the question, \emph{Is parsimony very important?} arises, which advises the use of deep learning models if this is the case. Deep learning models are designed to work with sequential data and do not need the aggregation sequence encoding as mentioned in Section~\ref{sec:preliminary}. Then, only when the response to the question: \emph{is the metric IRC unimportant?} is yes, the guidelines advise using CNN over LSTM. If, on the other hand, the response to the question \emph{Is faithfulness important?} is positive, the guidelines advise using the XGB model. It is necessary to mention that the faithfulness values for the GLRM are better (on average), but the predictive performance is better (on average) for the XGB model. Therefore, we take this trade-off into account. The next question, \emph{Is parsimony unimportant?} is meant to distinguish between RF and the LR (and advanced LLM). When parsimony is still essential, it is suggested to first use LR and/or LLM, depending on whether the data is heterogeneous or not. The RF model is recommended when you prefer models that use all the attributes. Next, the XGB model is advised when a low value for LOD is required. The LLM model should be preferred over the LR model when the data is heterogeneous. 

Note that in the case of BPIC2015(1) (see Section~\ref{sec:evaluation}), the LR and LLM were both more parsimonious, more faithful and obtained a higher AUC than RF. This means that, although the guidelines advise the use of GLRM (or DL models) to obtain interpretable models (based on parsimony), the LR was also very interpretable. This is due to the fact that we are working with aggregated values, so these guidelines serve rather as a guideline (and therefore should not be taken strictly). 
\section{Conclusion}\label{sec:conclusion}

This paper introduced a framework of metrics to evaluate the explainability of predictive models used for POP purposes. Compared to typical performance-based metrics and standard XAI metrics, they take into account the event, case, and control perspective and describe the \emph{interpretability} and the \emph{faithfulness}. We provide an extensive benchmark study of seven models and thirteen event logs, which illustrate how these metrics can capture different interpretations relevant to process data. Finally, we provide the reader with a consistent overview of the insights obtained by this study in the field of POP through a framework of guidelines contrasting traditional machine learning, deep learning and explainability approaches to guide the practitioner to the best model selection.
Not only can we conclude that the transparent GLRM model exhibits very interpretable explanations for high dimensional and sequential data (for only a small loss of performance), but also that machine learning models tend to perform better than the deep learning models based on predictive performance. Next, we also show that the interpretability-predictive performance trade-off holds in the field of POP when considering the introduced interpretability metric~\emph{parsimony}. To conclude, we show that each model has its advantages and disadvantages, where naively opting for the model with the highest performance can have a strong detrimental effect on both the interpretability and on the faithfulness of the explanations. As a result, identifying faithful explanations, while remaining interpretable, still imposes a challenge for black box models.

One of the limitations of this research is the fact that the metrics do not take the loss of information of the aggregation encoding into account. Therefore, it is assumed that the summary statistics of a certain variable contain no loss in information compared to the original attribute values. 
Future work will focus on the concept of Responsible AI, by introducing causal inference to assess the causal insights obtainable from predictive models, and focus on the creation of fairness-aware decision models.

\section*{Acknowledgments}
This was was supported in part by......

\bibliographystyle{unsrt}  
\bibliography{references}

\end{document}